Tech Science Press

# An Improved Time Feedforward Connections Recurrent Neural Networks

## Jin Wang [1,2], Yongsong Zou[1] and Se-Jung Lim[3,*]

[1] School of Hydraulic & Environmental Engineering, Changsha University of Science & Technology, Changsha, 410014, China

[2] School of Computer & Communication Engineering, Changsha University of Science & Technology, Changsha, 410014, China

[3] AI Liberal Arts Studies, Division of Convergence, Honam University, Gwangju-si, 62399, Korea

*Corresponding Author: Se-Jung Lim. Email: limsejung@honam.ac.kr

**Abstract:** Recurrent Neural Networks (RNNs) have been widely applied to deal with temporal problems, such as flood forecasting and financial data processing. On the one hand, traditional RNNs models amplify the gradient issue due to the strict time serial dependency, making it difficult to realize a long-term memory function. On the other hand, RNNs cells are highly complex, which will significantly increase computational complexity and cause waste of computational resources during model training. In this paper, an improved Time Feedforward Connections Recurrent Neural Networks (TFC-RNNs) model was first proposed to address the gradient issue. A parallel branch was introduced for the hidden state at time t-2 to be directly transferred to time t without the nonlinear transformation at time t-1. This is effective in improving the long-term dependence of RNNs. Then, a novel cell structure named Single Gate Recurrent Unit (SGRU) was presented. This cell structure can reduce the number of parameters for RNNs cell, consequently reducing the computational complexity. Next, applying SGRU to TFC-RNNs as a new TFC-SGRU model solves the above two difficulties. Finally, the performance of our proposed TFC-SGRU was verified through several experiments in terms of long-term memory and anti-interference capabilities. Experimental results demonstrated that our proposed TFC-SGRU model can capture helpful information with time step 1500 and effectively filter out the noise. The TFC-SGRU model accuracy is better than the LSTM and GRU models regarding language processing ability.

## 1 Introduction

Deep neural networks (DNNs) can be applied to learn the abstracted information of complex systems by constructing a deep model. They can be widely applied to pattern recognition[1-3], machine learning[4,5], etc.

Recurrent Neural Networks (RNNs) are the common DNNs used to handle time series tasks [6]. An RNN is different from other layer structures in hierarchical networks because of its time serial dependency in the same layer. It means the state $h^t$ at time t is determined by the state $h^{t-1}$ at time t-1. Due to the time serial, RNNs can capture the long-term dependency of the temporal data[7]. Thus, RNNs are the adequate technology for machine translation [8] and speech recognition [9,10]. Meanwhile, RNNs also play an essential role in specific video recognition, such as contextual video recognition [11] and visual sequence tasks [12,13].

However, training RNNs is difficult when long-term dependencies are large enough. The backpropagation chain of RNNs is longer during training than DNNs with the same number of layers, thus making it easy for vanishing and exploding gradients to arise [14]. When training RNNs, Backpropagation Through Time (BPTT) is used to update the parameters. The gradient problem encountered by RNNs results from continuous multiplication[15]. Researchers have put forward many improved models to improve these problems. The best solution is to introduce "gates" into the RNNs cell, such as Long Short-Term Memory (LSTM) [16,17]and Gate Recurrent Unit (GRU) [18].

The gates of LSTM and GRU alleviate the gradient vanishing and exploding with the cost of being time-consuming and having high computational complexity. In recent years, ResNet[19] and Highway Network[20] have been accepted to improve gradient problems in intense networks. These methods assume separate Feedforward Connections (FC) between different layers. Therefore, when gradient backpropagation is performed during training, it can be transferred to the previous network layer without nonlinear conversion. Recurrent Highway Network (RHN) reduces the cost of RNNs by feedforward connections between recurrent layers by introducing Highway Network [21]. But RHN cannot solve the gradient issues in the process of horizontal propagation in training. The residual recurrent networks (Res-RNN) use residual learning to improve the gradient problems[22]. Compared with Highway Network, ResNet cannot freely control the residual channels by learning gains. However, learning to open the information flow intelligently for a temporal network is helpful for long-term dependence.

Therefore, inspired by Highway Network, we proposed Time Feedforward Connections Recurrent Neural Networks (TFC-RNNs). TFC-RNNs improve RNNs by feedforward connections between the horizontal time steps, which allows the gradient at t-1 to be transferred to t+1 without the nonlinear transformation at time t. Meanwhile, to solve the overly complex problem of "gates" RNNs cell, a simplified RNNs cell model was proposed, that is, Simple Gate Unit (SGRU). Experimental results show that our TFC-SGRU model is superior to LSTM and GRU models in capturing long-term dependence.

The main contributions of this paper are summarized as follows. First, this paper introduced a novel recurrent unit with TFC-RNNs to solve the gradient issues. Then, to solve the overly complex problem of "gates" RNNs cell, a simplified RNNs cell model was proposed: SGRU. Finally, we have a combined model, TFC-SGRU, which improves gradient problems and reduces computational complexity.

The rest of this paper is organized as follows. Section 2 describes the relevant works. Section 3 presents the relevant TFC-RNNs and SGRU models. Section 4 provides extensive experimental results regarding Memory Copying Tasks, Denoise Tasks, and model accuracy of language processing ability based on bAbI Question Answering. Section 5 concludes this paper.

## 2 Related Work

### 2.1 Modified RNNs

The classical RNNs model is

$$h^t = f(Wx^t + Vh^{t-1} + b),\tag{1}$$

where $x^t$ and $h^t$ represent the input matrix and hidden state at time t, respectively. $h^{t-1}$ denotes the hidden state at time t-1. $f(\cdot)$ is the nonlinear transformation function. $W$ and $V$ are the input weight matrix and hidden output state weight matrix. $\boldsymbol{b}$ is the bias vector.

Despite the excellent performance of classical RNNs in processing time series data, there remain some problems: gradient issues and high computational complexity [23]. For DNN, there are many ways to improve the performance, such as transfer learning mentioned [24] and Faster R-CNN [25]. Some improved versions of RNNs have been proposed. Currently, LSTM and GRU are the best-performing RNNs of the improved structure. The expression of GRU is shown as follows.

$$z^t = \sigma(W_z x^t + V_z h^{t-1} + b_z),\tag{2}$$
$$r^t = \sigma(W_r x^t + V_r h^{t-1} + b_r),\tag{3}$$
$$\tilde{h}^t = tanh(W_h x^t + V_h(r^t \otimes h^{t-1})),\tag{4}$$
$$h^t = (1-z^t) \otimes h^{t-1} + z^t \otimes \tilde{h}^t,\tag{5}$$

where $r^t$ and $z^t$ represent the "reset gate" and "update gate" at time t, respectively. $h^t$ denotes hidden output at time t. $tanh$ is the activation function. And $tanh: y = \frac{e^x - e^{-x}}{e^x + e^{-x}}$. $W$ and $V$ are the input weight matrix and hidden output state weight matrix. $\boldsymbol{b}$ is the bias vector.

There are also many ways to handle temporal data, such as RNNs and federated learning[26]. RNNs remain irreplaceable in terms of temporal data, more and more widely used, such as weather forecasting [27-29], water conservancy prediction [30-32], etc. Therefore, the improved research on RNNs still plays a significant role. Besides, the new RNNs models are constantly emerging. Clockwork RNN (CW-RNN) is put forward to optimize long-term dependence ability in [33]. Through the design of a manner hiding memory matrix and a clock-like frequency mask, the RNN memory is divided into several parts to enhance the memory effects. Although such a method performs well in the regression model, it still has shortcomings in other data. In [34], Tree Memory Network (TMN) was proposed to achieve modular improvement for RNNs, with three modules defined: input module, controller, and memory module. Such a structure can capture the short-term and long-term dependence more efficiently. However, the input of prior knowledge is required, and the model performance is constrained. Despite the excellent performance of these improved models in specific fields, they can still not significantly improve the time serial dependency of RNNs, which denies the improvement outcome.

An Independently Recurrent Neural Network (IndRNN) [35]is proposed to optimize the computing complexity [35]. The cell of IndRNN in each layer is independent of each other, and the neuron in the next layer relates to all cells of the upper layer (the output of the upper layer is adopted as the input of a neuron in this layer). Breaking the series relationship of RNNs in the time dimension reduces the level of its complexity, but the long-term dependence ability of the model is reduced. The adder works similarly to RNNs [36]. Combined with the features of the adder, Carry-lookahead  (CL-RNNs) is proposed in [37], which alleviates the problems in time series for parallel computing, thus mitigating the effects of the training complexity. However, the performance of CL-RNN in time series tasks is inferior to LSTM and GRU.

Further, many researchers tried introducing residual learning into RNNs with residual connections proved effective for deep networks. RHN and Residual Recurrent Highway Network ((R2HN)[38] improve multi-layers RNNs by introducing Highway Network. However, there cannot improve the time serial dependency of RNNs.

Most RNNs are improved in certain aspects, so their overall performance remains inferior to LSTM and GRU. On the one hand, the TFC-SGRU model proposed in this paper enhances the time series by introducing the time feedforward. On the other hand, it improves the complexity of computing in training by reducing the cell complexity.

### 2.2 Gradient Analysis

In general, the performance of DNNs is related to the network depth in the spatial dimension. In the case of sufficient training data, the higher the depth, the greater the number of network layers, and the better the performance. Deep network training refers to the chain rules' reverse gradient update.

When there are N layers in one DNN and one neuron in each layer, it can be expressed as follows.

$$y_i = f(z_i), \tag{6}$$

$$z_i = w_i x_i + b_i, \tag{7}$$

where $y_i, x_i\, w_i, \boldsymbol{b_i}$ represent the output matrix, input matrix, weight matrix, and bias vector at the i ($i \in [1, N]$) layer, respectively. $f(\cdot)$ refers to the neuron activation function. When J denotes the loss of the DNN, the gradient of the weight matrix at the first layer can be calculated. The equation is presented as follows.

$$\frac{\partial J}{\partial w_1} = \frac{\partial J}{\partial y_N} \frac{\partial y_N}{\partial z_N} \frac{\partial z_1}{\partial w_1} \prod_1^N (\frac{\partial z_i}{\partial x_i} \frac{\partial x_i}{\partial z_{i-1}}), \tag{8}$$

$$\frac{\partial J}{\partial w_1} = \frac{\partial J}{\partial y_N} \prod_1^N \left( w_i f'(z) \right), \tag{9}$$

where $f'$ is the derivative of the activation function.

According to Eq. (9), the gradient transfer can be carried out through successive multiplication. At $|wf'(z)| < 1$, with the number of network layers increasing, the gradient will diminish continuously to approach 0, which hinders the network from parameters update. At $|wf'(z)| > 1$, the more layers of the network, the greater the gradient, which distorts the knowledge learned by the network.

For RNNs, the gradient problems are manifested in the time dimension. According to Eq. (1), its gradient update can be inferred.

$$\frac{dJ}{dV} = \sum_{t_1=1}^{T} \frac{dJ^{t_1}}{dV} = \sum_{t_1=1}^{T} \sum_{t_2=1}^{t_1} \frac{\partial J^{t_1}}{\partial h^{t_1}} \frac{\partial h^{t_1}}{\partial h^{t_2}} \frac{\partial h^{t_2}}{\partial V}, \tag{10}$$

where T represents the time step of RNNs' input matrix.

According to the BPTT, the gradient flows across all the time steps. Therefore, in Eq. (10), $\frac{\partial h^{t_1}}{\partial h^{t_2}}$ is the key to calculating gradient $\frac{dJ}{dV}$.

$$\frac{\partial h^{t_1}}{\partial h^{t_2}} = \prod_{t=t_1}^{t_2} \frac{\partial h^t}{\partial h^{t-1}} = \prod_{t=t_1}^{t_2} V^T \, diag[f'(h^{t-1})]. \tag{11}$$

Like Eq. (9), the gradient update of RNNs remains dependent on continuous multiplication. Therefore, the gradient updating in the time dimension for RNN is the same as that of general deep networks in the spatial dimension. The greater the time depth, the more likely it is for gradient explosion and gradient disappearance to occur.

### 2.3 Highway Network

DNNs could be expressed as follows.

$$y = H(x, W_H) \tag{12}$$

where $W_H$ represents the weight matrix, while x and y refer to the input and output matrix, respectively. $H(\cdot)$ denotes the activation function. Highway Network introduces the feedforward connection to the network, and it could be expressed as follows.

$$y = H(x, W_H)T(x, W_T) + x \cdot C(x, W_C), \tag{13}$$

where $T(x, W_T)$ and $C(x, W_C)$ represent the "carry gate" and "transform gate", respectively.

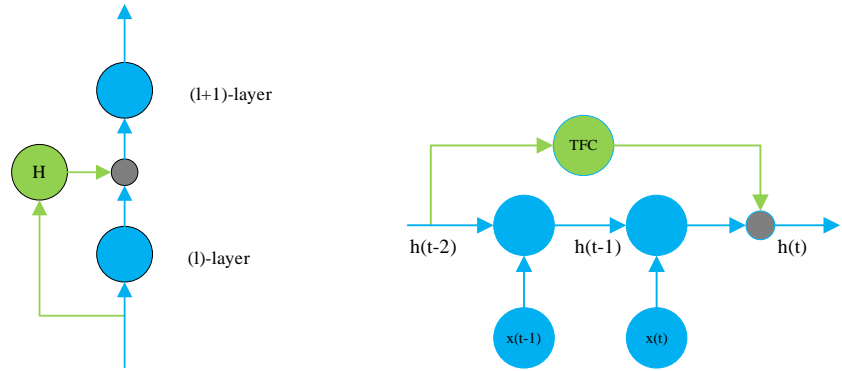

(a) The structure of the Highway network.          (b) The structure of TFC-RNNs

**Figure 1:** Comparison of the structure of Highway Network and RNNs.

The structure of the Highway Network is shown in Fig. 1(a). The green circles represent the FC. When the depth of the network is sufficient, the gradient will transfer the i+1 layer to the i-1 layer. As a result, the effective gradient will be transmitted further.

Concurrent with Highway Network, ResNets present residual connections with FC. Both are to learn high accuracy gains with enormously increased depth. Highway Network has parameters, and ResNets are parameter-free. Thus, Highway Network gates can freely control the channels by learning gains. Therefore, we use FC to solve the gradient issues in the process of horizontal propagation. However, RNNs need to control the flow of historical information intelligently, and we introduce Highway Network to improve RNNs. The structure of TFC-RNNs is shown in Fig. 1(b).

## 3 Our Proposed TFC-SGRU Model

### 3.1 TFC-RNNs Model

We present and explain the design of TFC-RNNs in this section. The RNNs input data typically consists of T time steps where the $t^{th}$ time step (t ∈ (1, 2, ..., T)) applies a nonlinear transform H on its input $x^t$ to produce its hidden output $y^t$. TFC-RNNs consist of two parts: TFC block and RNN cell, which can be seen in Fig. 2.

$$y^t = f(x^t, h^{t-1}), \tag{14}$$

where $f(\cdot)$ is usually an affine transform followed by a nonlinear activation function. $h^{t-1}$ represents the hidden state output at time t-1. $y^t$ is the hidden output of RNNs at time t, shown in Fig. 2 RNN cell block.

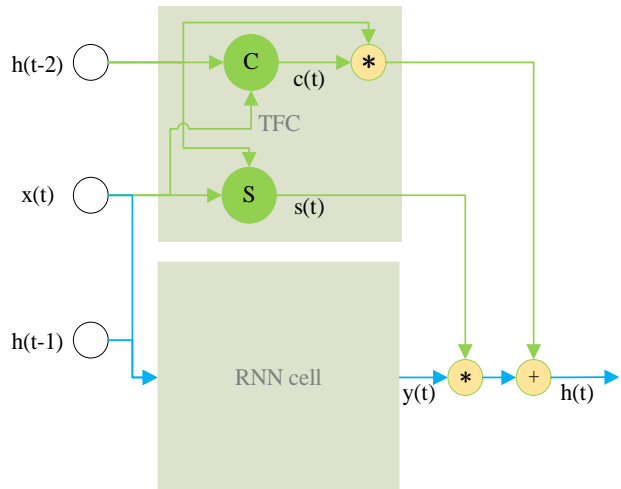

**Figure 2:** The TFC-RNNs cell.

We define two transforms, "carry gate" and "transform gate", such that

$$h^t = y^t \cdot s^t + h^{t-2} \cdot (1 - s^t), \tag{15}$$

where $s^t$ and $(1 - s^t)$ represent the "carry gate" and "transform gate", respectively. The gates can be seen in Fig. 2 TFC block. $h^t, h^{t-2}$ represent the hidden state output of TFC-RNNs at time t and t-2, respectively.

$$s^t = f(x^t, h^{t-2}), \tag{16}$$

where H is usually an affine transform followed by a nonlinear activation function.

If $s^t = 0$, the state $h^{t-2}$ can be directly transmitted to state $h^t$. Thus, the gradient at time t-2 could be directly transmitted to t without the transmission at time t-1, relieving the gradient problem. Since the gradient could be transmitted across the time step, RNNs can achieve more prolonged time independence.

### *3.2 Analysis of TFC-RNNs Gradient*

The gradient transmission of TFC-RNNs is analyzed to demonstrate the benefits of TFC to RNNs from a theoretical perspective. TFC-RNNs could be expressed as:

$$h^t = f(Wx^t + Vh^{t-1} + W_Hx^t + V_Hh^{t-2}),  \tag{17}$$

where $W_H$ and $V_H$ represent the TFC input weight at time t and TFC hidden state weight at time t-2, respectively. If losses of the network are J, the gradient of V could be calculated as:

$$\frac{dJ}{dV} = \sum_{t=1}^{T} \frac{dJ^t}{dV} = \sum_{t=1}^{T} \sum_{k=1}^{t} \frac{\partial J^t}{\partial h^t} \left( \frac{\partial h^t}{\partial h^k} \frac{\partial h^k}{\partial V} + \frac{\partial h^t}{\partial h^{k-1}} \frac{\partial h^{k-1}}{\partial V} \right).  \tag{18}$$

According to the chain rules, $\frac{\partial h^t}{\partial h^k}$ and $\frac{\partial h^t}{\partial h^{k-1}}$ are the keys to $\frac{dJ}{dV}$.

$$\frac{\partial h^t}{\partial h^k} = \frac{\partial h^t}{\partial h^{t-1}} \frac{\partial h^{t-1}}{\partial h^{t-2}} \cdots \frac{\partial h^{k+1}}{\partial h^k} = \prod_{j=k+1}^{t} \frac{\partial h^j}{\partial h^{j-1}} = \prod_{j=k+1}^{t} V^T diag[f'(h^{j-1})],$$

$$\frac{\partial h^t}{\partial h^{k-1}} = \frac{\partial h^t}{\partial h^{t-2}} \frac{\partial h^{t-2}}{\partial h^{t-4}} \cdots \frac{\partial h^{k+2}}{\partial h^k} = \prod_{j=k+1}^{t} \frac{\partial h^j}{\partial h^{j-2}} = \prod_{j=k+1}^{t} V^T diag[f'(h^{j-2})],  \tag{19}$$

where $V^T$ is V transpose. Substitute Eq. (19) into Eq. (18).

$$\frac{dJ}{dV} = \sum_{t=1}^{T} \sum_{k=1}^{t} \frac{\partial J^t}{\partial h^t} \left( \prod_{j=k+1}^{t} V^T diag[f'(h^{j-1})] \frac{\partial h^k}{\partial V} + \prod_{j=k+1}^{t} V^T diag[f'(h^{j-2})] \frac{\partial h^{k-1}}{\partial V} \right).  \tag{20}$$

The TFC module transforms the gradient into a sum expression instead of continuous multiplication, which avoids gradient issues. Besides, the TFC is trained and activated. The activation function can regularize the result in a perfect range.

### *3.3 Single Gate Recurrent Unit*

LSTM and GRU perform better in the time series data with increasing the number of parameters of RNNs cell. Miscellaneous leads to a significant increase in the number of training parameters, thus increasing the difficulty of training considerably. For DNN, there are many ways to reduce the computation, such as transfer learning mentioned in [24]. This suggests a simple RNNs cell that reduces the training parameters. This paper simplified the GRU cell to propose the more superficial RNNs cell, Single Gate Recurrent Unit (SGRU). SGRU can be expressed as follows.

$$r^t = \sigma(W_r x^t + V_r h^{t-1} + b_r),  \tag{21}$$

$$\bar{h}^t = tanh(Wx^t + V(r^t \otimes h^{t-1})),  \tag{22}$$

$$h^t = (1 - r^t) \otimes h^{t-1} + r^t \otimes \bar{h}^t,  \tag{23}$$

where $r^t$ and $h^t$ represent the "reset gate" and the hidden output state at time t, respectively. $\boldsymbol{b_r}$ indicates the bias vector. $W$ and V denote the weight matrix. $\sigma(\cdot)$ refers to the nonlinear function sigmoid, and $\sigma: y(x) = \frac{1}{1+e^{-x}}$. $tanh(\cdot)$ is the hyperbolic tangent function, and $tanh: y(x) = \frac{e^x - e^{-x}}{e^x + e^{-x}}$.

GRU relies on a "reset gate" to filter the noise in the history, and an "update gate" is applied to retain useful historical information, thus achieving long-term memory[39]. Compared with the GRU Eq. (2)-(5), SGRU integrates the "update gate" and "reset gate" to " reset gate" to reduce the number of its training parameters.

### *3.4 TFC-SGRU Model*

For training a network, we show the parameter for each part of TFC-RNNs. The "carry gate" can define as

$$s^t = \sigma(W_H x^t + V_H h^{t-2} + b_H),  \tag{24}$$

where $W_H$ and $V_H$ are both the weight matrix and $b_H$ the bias vector for the transform gates.

We found that a simple RNNs cell is sufficient for learning in long-time dependencies networks for different activation functions. This is a significant property since it may not be possible to reduce the number of training parameters. RNN cell block can be set to our proposed SGRU model. Finally, we have a combined model, namely the TFC-SGRU model.

## 4 Experimental Results and Analysis

We compare TFC-SGRU with the well-known RNNs (LSTMs and GRUs). Our code can be downloaded from https://gitee.com/homesong/tfcrnn.

### 4.1 Memory Copying Tasks

Memory copying tasks are the benchmark to verify a model's performance in capturing long-term dependencies, which follows a similar configuration. A data table consists of 10 categories: ( $a^i$ ), and i ∈ [0,9], ( $a^i$ )$_0^7$ represent data. $a^8$ and $a^9$ represent "blank" and "marker", which are separators. The input takes the form of a T+20 length vector of categories. The first ten characters are sampled from ($a^i$)$_0^7$, and the next T-1 characters are $a^8$. The next is a character that is set to $a^9$. Also, $a^9$ is a separator. The remaining ten characters are set to $a^8$. The expected output is comprised of T + 10 characters of $a^8$ and ten characters of the input sequence in the same order. The cost function is cross-entropy, as follows

$$C = (10 * \log(8))/(T + 20). \tag{25}$$

**Table 1:** Copying tasks parameters

| Parameters | Value |
|---|---|
| Learning rate | 0.001 |
| Decay rate | 0.9 |
| Batch size | 128 |

This task is used to train the network to remember the ten categories and the order of characters for the T time step. A simple baseline is established to output 10+T blanks followed by ten random symbols, and a cross-entropy is produced as follows.

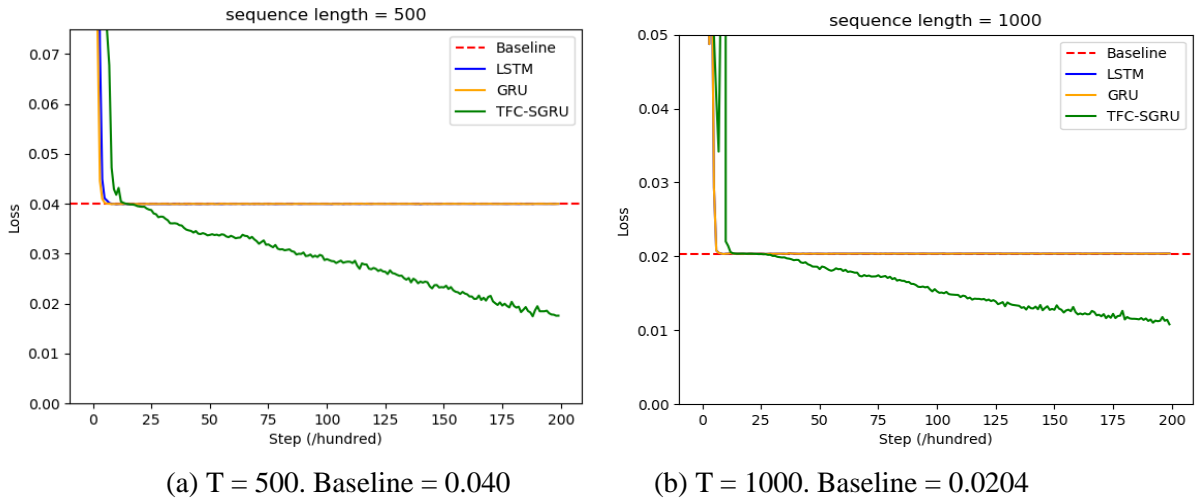

(a) T = 500. Baseline = 0.040                    (b) T = 1000. Baseline = 0.0204

**Figure 3:** Result of copying tasks. These show the cross-entropy loss of training.

This experiment is aimed mainly to verify the performance of TFC-SGRU in long-term dependence. The performance was tested at time step t=500 and 1000, respectively. Allowing for the activation function of the hidden output, tanh was adopted. And the optimizer is set to RMSProp. The control model includes the widely recognized LSTM and GRU models. Other model hyperparameters were set according to Tab. 1.

The cross-entropy loss gets stuck, indicating that the gradient disappears and that the model's parameter cannot be updated. According to Eq. (11), the more significant time step T increases, the easier it is for the gradient to disappear and the more difficult it is to capture long-term dependence. Fig. 3(a) shows the result of the copying task for T =500. The baseline is 0.04. TFC-SGU is the only structure to break down the baseline successfully, while GRU and LSTM get stuck. The TFC-SGU outperforms the LSTM and GRU in terms of learnability. It indicates that it is easy for TFC-SGU to learn the knowledge before time step 500. To further increase the sequence length, T is set to 1000. The baseline is 0.0204. Fig. 3(b) shows the result of the copying task. LSTM and GRU fall outside the baseline, while TFC-SGRU could break the baseline easily. Therefore, the capacity of TFC-SGRU to capture the long-term dependence is much higher than LSTM and GRU.

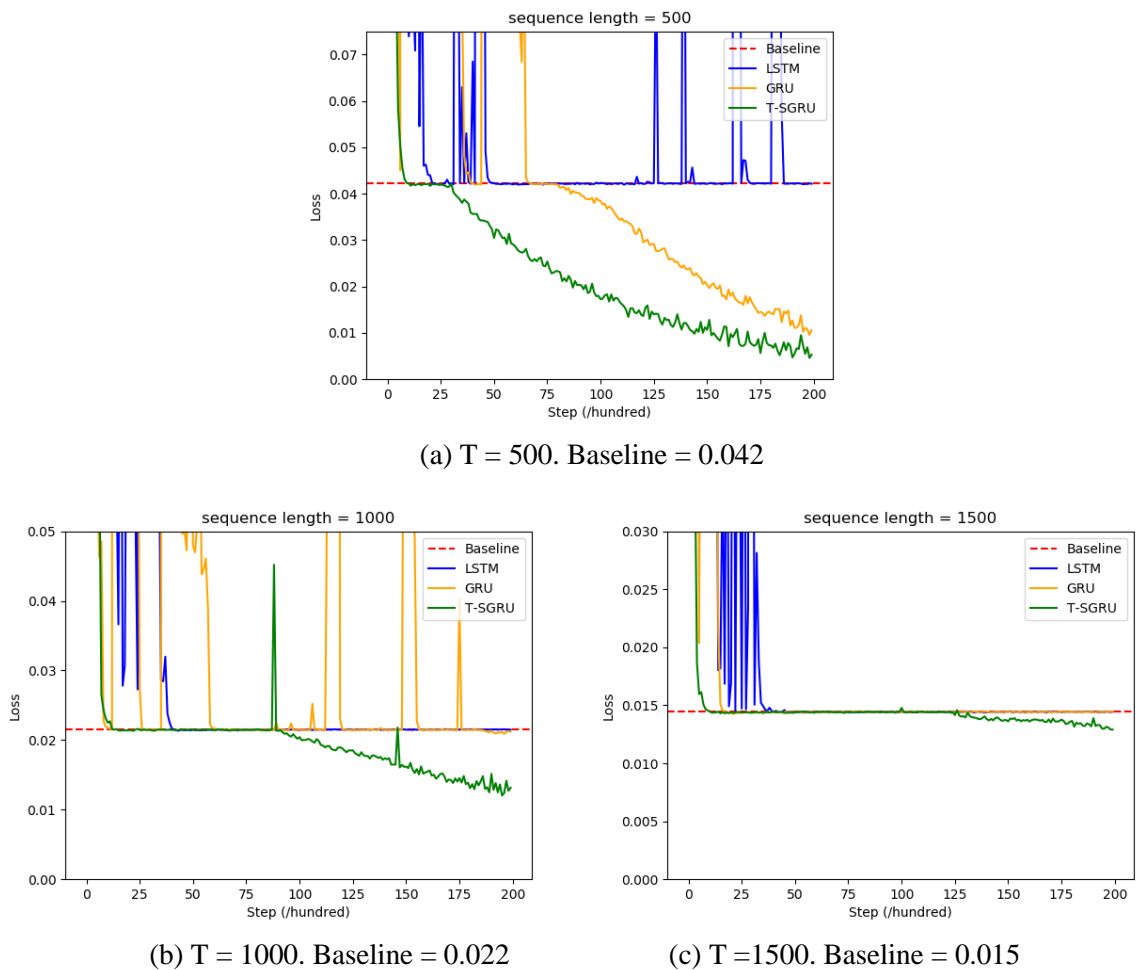

(a) T = 500. Baseline = 0.042

(b) T = 1000. Baseline = 0.022                    (c) T =1500. Baseline = 0.015

**Figure 4:** Results of the denoise tasks. These show the cross-entropy loss of training.

### 4.2 Denoise Tasks

Denoise tasks are used to evaluate the capability of RNN architecture to remove noise (forgetting ability). The random noise is added into sequential labels, while RNN locates the noise and outputs the sequential labels. The composition of the denoise tasks data is similar to that of copying tasks. An alphabet consists of i categories $(a^i)$, and i ∈ $(0,1,2 …, n − 1, n)$. $a^{n−1}$ and $a$ represents "noise" and "marker", respectively. In this paper, n is set to 10. Showing similarity to the task, the baseline for this task outputs 10+T blanks followed by ten random symbols and produces a cross-entropy.

$$C = (10 * \log(9))/(T + 20). \tag{26}$$

The experiment was performed to verify the capacity of TFC-SGRU to remove the historical noise. The performance was tested at time step t=500 and 1000, respectively. The other parameters were set in the same way as in Tab. 1. The optimizer is also set to RMSProp. The control model remains LSTM and GRU. Fig. 4(a) shows the experimental results obtained when the time step is T=500. LSTM can break the baseline of 0.04. Although TFC-SGRU and GRU could break the baseline effectively, the speed of TFC-SGRU is significantly faster than GRU. TFC-SGRU and GRU effectively remove long-term noise beyond the time step T=500. By contrast, LSTM is ineffective. Fig. 4(b) and 4(c) show the experimental results at time steps T=1000 and T=1500. In two experimental results, only TFC-SGRU effectively broke the baseline, indicating that the model remembers the helpful information before the time steps 1500 and filters the long-term noise.

### 4.3 Question Answering

We use the bAbI Question Answering datasets[40] to examine RNN's ability to understand language. bAbI datasets include 20 different subtasks. These tasks are to find the answer to the question according to the story. A story consists of several sentences. The answers are a word or several words. Every different task includes 1k/1k/9k examples for test/dev/train. Each subtask is trained separately. The input data is $(x_1^s, x_2^s, x_3^s, \dots x_n^s, x^q)$, where s is the story, q is the question, and n is the number of sentences for the story.

**Table 2:** The test accuracy (%) on the bAbI dataset.

| ID | Task | TFC-SGRU | LSTM | GRU | Baseline[40] |
|----|------|----------|------|-----|--------------|
| 1 | Single Supporting Fact | **74.1** | 51.3 | 50.5 | 50.0 |
| 2 | Two Supporting Facts | 41.6 | 42.3 | 42.1 | 20.0 |
| 3 | Three Supporting Facts | 45.2 | 48.7 | 37 | 20.0 |
| 4 | Two Arg. Relations | 68.4 | 66.6 | 64.2 | 61.0 |
| 5 | Three Arg. Relations | 83.5 | 83.5 | 84.2 | 70.0 |
| 6 | Yes/No Questions | 73.1 | 50.3 | 68.5 | 48.0 |
| 7 | Counting | 79.6 | 80.4 | 79.5 | 49.0 |
| 8 | Lists/Sets | 89.4 | 77.9 | 90.5 | 45.0 |
| 9 | Simple Negation | 63.8 | 63.8 | 74.7 | 64.0 |
| 10 | Indefinite Knowledge | 57.4 | 66.7 | 60.9 | 44.0 |
| 11 | Basic Coreference | 87.7 | 85.2 | 74.6 | 72.0 |
| 12 | Conjunction | **93.1** | 94.1 | 76.8 | 74.0 |
| 13 | Compound Coref | 94.4 | 94.3 | 94.4 | 94.0 |
| 14 | Time Reasoning | 45.7 | 48.5 | 39.7 | 27.0 |
| 15 | Basic Deduction | 60.9 | 58.7 | 68.4 | 21.0 |
| 16 | Basic Induction | 48.2 | 50.4 | 48 | 23.0 |
| 17 | Positional Reasoning | 59.7 | 48.0 | 59.7 | 51.0 |
| 18 | Size Reasoning | 46.9 | 58.5 | 53.1 | 52.0 |
| 19 | Path Finding | **18.1** | 9.8 | 8.3 | 8.0 |
| 20 | Agent's Motivations | 98.2 | 98.3 | 98.9 | 91.0 |
| | Mean Accuracy(%) | 66.45 | 63.87 | 63.70 | 49.20 |

This experiment tests the GRU, LSTM, and TFC-SGU on sentence-level bAbI datasets. We use Adam optimizer to train the model with a batch size of 128, a hidden state size of 40, layers of 2, and embeddings of 128. The learning rate can be set to 0.001. The characteristics of each subtask are different, so RNNS accuracy is different on 20 subtasks. Tab. 2 shows the accuracy result on TFC-SGRU, LSTM, and GRU. Given the same parameters, in most subtasks, the accuracy of TFC-SGRU is better than that of LSTM/GRU. TFC-SGRU achieves the highest average accuracy of 20 subtasks.

## 5 Conclusion

In this paper, we firstly proposed TFC-RNN by introducing the Time Feedforward Connections. The TFC-RNNs model can transmit the gradient from time t-2 to t without t-1 nonlinear transformation, improving the long-term dependencies and the gradient problem. Then, we proposed SGU by removing the update gate of GRU. TFC-RNNs and SGRU were combined as a new TFC-SGRU model to reduce computational complexity. Finally, our experimental results show that optimization of TFC-SGRU is not hampered even as time-depth increases to 1500-time steps. However, TFC-SGRU still has limitations in parallel computing. Next, we will explore efficient RNNs sequence models.

**Acknowledgement:** This work was funded by the National Science Foundation of Hunan Province (2020JJ2029). This work was also supported by a research fund from Honam University, 2022.

**Conflicts of Interest:** The authors declare no conflicts of interest to report regarding the present study.